\documentclass[runningheads]{llncs}
\usepackage[T1]{fontenc}
\usepackage{graphicx}
\usepackage{hyperref}
\usepackage{color}

\usepackage{amsmath,amsfonts,bm}

\usepackage{subcaption}
\usepackage{tikz}
\usetikzlibrary{shapes,arrows,backgrounds,fit,positioning,calc,angles,quotes,arrows.meta,trees,shapes.geometric,shapes.symbols,decorations.pathreplacing,decorations.pathmorphing,patterns}
\usepackage{tikz-qtree}

\makeatletter
\pgfdeclareshape{document}{
\inheritsavedanchors[from=rectangle] % this is nearly a rectangle
\inheritanchorborder[from=rectangle]
\inheritanchor[from=rectangle]{center}
\inheritanchor[from=rectangle]{north}
\inheritanchor[from=rectangle]{south}
\inheritanchor[from=rectangle]{west}
\inheritanchor[from=rectangle]{east}
\backgroundpath{% this is new
\southwest \pgf@xa=\pgf@x \pgf@ya=\pgf@y
\northeast \pgf@xb=\pgf@x \pgf@yb=\pgf@y
\pgf@xc=\pgf@xb \advance\pgf@xc by-10pt % this should be a parameter
\pgf@yc=\pgf@yb \advance\pgf@yc by-10pt
\pgfpathmoveto{\pgfpoint{\pgf@xa}{\pgf@ya}}
\pgfpathlineto{\pgfpoint{\pgf@xa}{\pgf@yb}}
\pgfpathlineto{\pgfpoint{\pgf@xc}{\pgf@yb}}
\pgfpathlineto{\pgfpoint{\pgf@xb}{\pgf@yc}}
\pgfpathlineto{\pgfpoint{\pgf@xb}{\pgf@ya}}
\pgfpathclose
\pgfpathmoveto{\pgfpoint{\pgf@xc}{\pgf@yb}}
\pgfpathlineto{\pgfpoint{\pgf@xc}{\pgf@yc}}
\pgfpathlineto{\pgfpoint{\pgf@xb}{\pgf@yc}}
\pgfpathlineto{\pgfpoint{\pgf@xc}{\pgf@yc}}
}
}
\makeatother

\newcommand{\refassumption}[1]{Assumption~\ref{#1}}

\newcommand{\reffigure}[1]{\figurename~\ref{#1}}

\newcommand{\refsection}[1]{Section~\ref{#1}}
\newcommand{\refsubsection}[1]{Subsection~\ref{#1}}

\newtheorem{assumption}{Assumption}

\usepackage[skins,breakable,xparse]{tcolorbox}
\tikzset{
  dirtree/.style={
    grow via three points={one child at (0.8,-0.7) and two children at (0.8,-0.7) and (0.8,-1.45)}, 
    edge from parent path={($(\tikzparentnode\tikzparentanchor)+(.4cm,0cm)$) |- (\tikzchildnode\tikzchildanchor)}, growth parent anchor=west, parent anchor=south west},
}

\usepackage{listingsutf8}

\lstdefinelanguage{ludii}{
  keywords={aPI,card,component,die,piece,domino,path,tile,board,boardless,surakartaBoard,hashiBoard,puzzleBoard,track,container,deck,dice,hand,equipment,item,dominoes,hints,map,regions,all,allDiceEqual,allDiceUsed,allPassed,allType,allClaimed,baseBooleanFunction,booleanConstant,booleanFunction,can,canMove,canType,all,allDifferent,allPuzzleType,forAll,isUnique,is,isPuzzleGraphType,isPuzzleRegionResultType,isPuzzleSimpleType,isCount,isSum,allConnected,isProduct,isShape,sameParity,isSolved,isThreatened,isWithin,isBlocked,isConnected,isCrossing,isLastFrom,isLastTo,isIn,isInvisible,isMasked,isVisible,isAnyDie,isEven,isFlat,isOdd,isPipsMatch,isSidesMatch,isVisited,is,isComponentType,isConnectType,isEdgeType,isGraphType,isIndexPlayerType,isIntegerType,isInType,isLineType,isLoopType,isPathType,isPlayerType,isRegularGraphType,isRelationType,isSimpleType,isSiteType,isStringType,isTargetType,isTreeType,isLine,isLoop,isPath,isEnemy,isFriend,isMover,isNext,isPrev,isTriggered,isRegularGraph,isRelated,isCycle,isFull,isPending,isEmpty,isOccupied,isDecided,isProposed,isTarget,isCaterpillerTree,isSpanningTree,isTree,isTreeCentre,isPlanarGraph,isPlanarGraph2,and,equals,ge,gt,if,le,lt,not,notEqual,or,xor,no,noMoves,noType,was,wasPass,wasType,directions,directionsFunction,if,baseFloatFunction,floatConstant,floatFunction,baseGraphFunction,basis,brick,brickShapeType,diamondOrPrismOnBrick,spiralOnBrick,squareOrRectangleOnBrick,celtic,customOnHex,diamondOnHex,hex,hexagonOnHex,hexShapeType,rectangleOnHex,starOnHex,triangleOnHex,customOnMesh,mesh,morris,quadhex,customOnSquare,diagonalsType,diamondOnSquare,rectangleOnSquare,square,squareShapeType,tiling,tiling31212,tiling333333_33434,tiling33336,customOn33344,tiling33344,tiling33434,customOn3464,diamondOn3464,hexagonOn3464,parallelogramOn3464,rectangleOn3464,starOn3464,tiling3464,tiling3464ShapeType,triangleOn3464,customOn3636,tiling3636,tiling4612,customOn488,squareOrRectangleOn488,tiling488,tilingType,customOnTri,diamondOnTri,hexagonOnTri,rectangleOnTri,starOnTri,tri,triangleOnTri,triShapeType,circle,rectangle,repeat,shape,shapeTypeStar,spiral,wedge,graphFunction,add,clip,complete,dual,hole,intersect,keep,makeFaces,merge,remove,renumber,rotate,scale,shift,skew,splitCrossings,subdivide,trim,union,baseIntArrayFunction,intArrayConstant,intArrayFunction,difference,degrees,groupSizes,rotations,baseIntFunction,ahead,centrePoint,column,coord,cost,handSite,id,layer,mapEntry,phase,row,where,card,cardSimpleType,cardSiteType,cardTrumpSuit,cardRank,cardSuit,cardTrumpRank,cardTrumpValue,groupProduct,tree,between,edge,from,hint,level,site,to,track,var,countPieces,countPips,count,countComponentType,countSimpleType,countSiteType,countSpaceType,countStepsType,countActive,countCells,countColumns,countEdges,countMoves,countMovesThisTurn,countPhases,countPlayers,countRows,countTrials,countTurns,countVertices,countAdjacent,countDiagonal,countNeighbours,countNumber,countOff,countOrthogonal,countSites,countGroups,countLiberties,countSteps,countDegreeProduct,countLeaves,countNonLeaves,countNonLeavesDegree,countTreeDepth,countTreeSize,face,pips,intConstant,intFunction,last,lastFrom,lastTo,lastType,matchScore,abs,add,div,if,max,min,mod,mul,pow,sub,sizeGroup,sizeTerritory,sizeStack,size,sizeGroupType,sizeSiteType,sizeTerritoryType,topLevel,amount,counter,mover,next,previous,score,state,what,who,pathExtent,trackSiteMove,trackSiteEndTrack,trackSite,trackSiteMoveType,trackSiteType,valuePiece,valuePlayer,valuePending,value,valueComponentType,valuePlayerType,valueSimpleType,tree,triangleGroupCount,baseRangeFunction,exact,max,min,range,rangeFunction,baseRegionFunction,forEach,difference,expand,if,intersection,union,regionConstant,regionFunction,sitesAround,sitesCoords,sitesCrossing,sitesCustom,sitesDirection,sitesDistance,sitesAngled,sitesAxial,sitesHorizontal,sitesSlash,sitesSlosh,sitesVertical,sitesGroup,sitesIncident,sitesCell,sitesColumn,sitesEdge,sitesEmpty,sitesPhase,sitesRow,sitesState,sitesLineOfSight,sitesOccupied,sitesEquipmentRegion,sitesHand,sitesInvisible,sitesMasked,sitesStart,sitesTrack,sitesVisible,sitesWinning,sitesSide,sitesBoard,sitesBottom,sitesCentre,sitesConcaveCorners,sitesConvexCorners,sitesCorners,sitesHint,sitesInner,sitesLastFrom,sitesLastTo,sitesLeft,sitesLineOfPlay,sitesMajor,sitesMinor,sitesOuter,sitesPending,sitesPlayable,sitesRight,sitesToClear,sitesTop,sites,sitesAroundType,sitesCrossingType,sitesDirectionType,sitesDistanceType,sitesEdgeType,sitesGroupType,sitesIncidentType,sitesIndexType,sitesLineOfSightType,sitesOccupiedType,sitesPlayerType,sitesSideType,sitesSimpleType,sitesTrackType,sitesWalk,borderSites,game,games,match,subgame,mode,player,players,baseEndRule,byScore,end,endRule,forEach,if,result,automove,meta,metaRule,noRepeat,swap,nextPhase,phase,baseMoves,decisions,move,moveBetType,moveFromToType,moveHopType,moveLeapType,moveMessageType,movePromoteType,moveRemoveType,moveSelectType,moveSetType,moveShootType,moveSimpleType,moveSiteType,moveSlideType,moveStepType,add,apply,bet,claim,effects,fromTo,hop,leap,note,attract,custodial,deal,directionCapture,enclosed,flip,allCombinations,and,append,if,or,push,roll,setDirection,setCounter,setVar,setNextPlayer,setPending,setScore,setValue,set,setDirectionType,setNextPlayerType,setPendingType,setPlayerType,setRegionType,setSiteType,setTrumpType,setValueType,setCount,setState,setInvisible,setMasked,setVisible,setTrumpSuit,sow,addScore,moveAgain,rememberState,shiftPlayers,swapPlayers,swapPieces,swap,swapPlayersType,swapSitesType,surrounded,takeControl,takeDomino,take,takeControlType,takeSimpleType,then,trigger,pass,playCard,promote,propose,remove,avoidStoredState,do,firstMoveOnTrack,maxDistance,max,maxDistanceType,maxMovesType,maxCaptures,maxMoves,priority,satisfy,select,shoot,slide,step,vote,forEachDie,forEachDirection,forEach,forEachDieType,forEachDirectionType,forEachPieceType,forEachPlayerType,forEachSiteType,forEachPiece,forEachPlayer,forEachSite,moves,play,rule,rules,deal,set,placeItem,place,placeRandomType,placeStackType,placeRandom,placeCustomStack,placeMonotonousStack,setAmount,setScore,setTeam,set,setStartGraphType,setStartPlayersType,setStartPlayerType,setStartSitesType,allInvisible,setCost,setCount,setPhase,setSite,split,splitType,start,startRule,basisType,landmarkType,puzzleElementType,regionTypeDynamic,regionTypeStatic,relationType,shapeType,siteType,stepType,tilingBoardlessType,trackType,cardType,dealableType,suitType,dummy,modeType,repetitionType,resultType,roleType,whenType,gameType,noType,absoluteDirection,compassDirection,direction,directionType,directionUniqueName,relativeDirection,rotationalDirection,spatialDirection,dummy,score,card,hint,region,values,edge,face,graph,graphElement,itemScore,measureGraph,perimeter,poly,properties,radial,radials,step,steps,trajectories,vertex,count,pair,between,flips,from,to,what,who,ai,aIItem,features,featureSet,heuristics,centreProximity,cornerProximity,currentMoverHeuristic,heuristicTerm,lineCompletionHeuristic,material,mobilitySimple,ownRegionsCount,playerRegionsProximity,playerSiteMapCount,regionProximity,score,sidesProximity,divNumBoardSites,divNumInitPlacement,heuristicTransformation,logisticFunction,tanh,bestAgent,pair,board,boardBooleanType,boardColourType,boardShapeType,boardStyleThicknessType,boardStyleType,boardCheckered,boardColour,boardShape,boardStyle,boardStyleThickness,graphics,graphicsItem,noAnimation,noBoard,noCurves,noHandScale,noMaskedColour,no,noBooleanType,adversarialPuzzle,stackType,suitRanking,pieceColour,pieceColourFromState,pieceFamilies,pieceBackground,pieceForeground,pieceAddStateToName,pieceExtendName,pieceRename,piece,pieceColourFromStateType,pieceColourType,pieceFamiliesType,pieceGroundType,pieceNameType,pieceReflectType,pieceScaleByType,pieceScaleType,pieceStyleType,pieceReflect,pieceScale,pieceScaleByValue,pieceStyle,playerColour,player,playerColourType,regionColour,region,regionColourType,showCost,showPits,showPlayerHoles,showRegionOwner,showCheck,showPieceState,showPieceValue,showEdges,showScore,show,showBooleanType,showCheckType,showComponentDataType,showComponentType,showEdgeType,showScoreType,showSiteDataType,showSiteType,showSymbolType,showSitesAsHoles,showSitesShape,showSymbol,boardGraphicsType,colour,colourRoutines,userColourType,componentStyleType,containerStyleType,controllerType,edgeInfoGUI,edgeType,lineStyle,metadataFunctions,metadataImageInfo,pieceStackType,valueLocationType,whenScoreType,aliases,author,classification,credit,date,description,origin,publisher,rules,source,version,info,infoItem,metadata,metadataItem},
  basewidth  = {.6em,0.6em},
  keywordstyle=\color{mblue}\bfseries,
  ndkeywords={Off,End,Undefined,DiceUsed,DiceEqual,Passed,Move,Different,Unique,Count,Sum,Solved,Threatened,Within,Connected,Blocked,Crossing,LastFrom,LastTo,Visible,Masked,Invisible,Odd,Even,Visited,SidesMatch,PipsMatch,Flat,AnyDie,In,Line,Loop,Path,Mover,Next,Prev,Friend,Enemy,Triggered,RegularGraph,Related,Cycle,Pending,Full,Empty,Occupied,Proposed,Decided,Target,Tree,SpanningTree,CaterpillerTree,TreeCentre,Moves,Pass,Square,Rectangle,Diamond,Prism,Spiral,Limping,NoShape,Square,Rectangle,Diamond,Triangle,Hexagon,Star,Limping,Prism,Implied,Solid,Alternating,Concentric,Radiating,NoShape,Square,Rectangle,Diamond,Limping,Custom,Square,Rectangle,Diamond,Prism,Triangle,Hexagon,Star,Limping,T31212,T3464,T488,T33434,T33336,T33344,T3636,T4612,T333333_33434,NoShape,Square,Rectangle,Diamond,Triangle,Hexagon,Star,Limping,Prism,Star,TrumpSuit,Rank,Suit,TrumpValue,TrumpRank,Pieces,Pips,Rows,Columns,Turns,Moves,Trials,MovesThisTurn,Phases,Vertices,Edges,Cells,Players,Active,Sites,Adjacent,Neighbours,Orthogonal,Diagonal,Off,Groups,Liberties,Steps,To,From,Group,Stack,Territory,Move,EndSite,Piece,Player,Pending,Around,Crossing,Direction,Distance,Axial,Horizontal,Vertical,Angled,Slash,Slosh,Group,Incident,Row,Column,Phase,Cell,Edge,State,Empty,LineOfSight,Occupied,Hand,Start,Track,Winning,Visible,Masked,Invisible,Side,Board,Top,Bottom,Left,Right,Inner,Outer,Corners,ConcaveCorners,ConvexCorners,Major,Minor,Centre,Hint,ToClear,LineOfPlay,Pending,Playable,LastTo,LastFrom,Track,Bet,FromTo,Hop,Leap,Propose,Vote,Promote,Remove,Select,Set,Shoot,Pass,PlayCard,Add,Claim,Slide,Step,Direction,NextPlayer,Pending,Value,Score,Visible,Masked,Invisible,Count,State,TrumpSuit,Counter,Var,Players,Pieces,Control,Domino,Distance,Moves,Captures,Die,Direction,Piece,Player,Site,Random,Stack,AllInvisible,Team,Amount,Score,Count,Cost,Phase,Deck,NoBasis,Triangular,Square,Hexagonal,T33336,T33344,T33434,T3464,T3636,T4612,T488,T31212,T333333_33434,SquarePyramidal,HexagonalPyramidal,Wheel,Circle,Spiral,Dual,Brick,Mesh,Morris,Celtic,QuadHex,CentreSite,LeftSite,RightSite,Topsite,BottomSite,FirstSite,LastSite,Cell,Edge,Vertex,Hint,Empty,NotEmpty,Own,NotOwn,Enemy,NotEnemy,AllPlayers,Rows,Columns,AllDirections,HintRegions,Layers,Diagonals,SubGrids,Regions,Vertices,Corners,Sides,SidesNoCorners,AllSites,Touching,Orthogonal,Diagonal,Off,Adjacent,All,NoShape,Custom,Square,Rectangle,Triangle,Hexagon,Cross,Diamond,Prism,Quadrilateral,Rhombus,Wheel,Circle,Spiral,Wedge,Star,Limping,Polygon,Vertex,Edge,Cell,F,B,L,R,Square,Triangular,Hexagonal,Track,Joker,Ace,Two,Three,Four,Five,Six,Seven,Eight,Nine,Ten,Jack,Queen,King,Dominoes,Cards,Clubs,Spades,Diamonds,Hearts,Alternating,Simultaneous,InTurn,InGame,Positional,Situational,Infinite,Win,Loss,Draw,Tie,Abandon,Crash,Neutral,P1,P2,P3,P4,P5,P6,P7,P8,P9,P10,P11,P12,P13,P14,P15,P16,Team1,Team2,Team3,Team4,Team5,Team6,Team7,Team8,Team9,Team10,Team11,Team12,Team13,Team14,Team15,Team16,Each,Shared,All,Any,Mover,Next,Prev,NonMover,Enemy,Ally,NonAlly,Partner,NonPartner,NonNeutral,Player,StartOfMove,EndOfMove,StartOfTurn,EndOfTurn,StartOfRound,EndOfRound,StartOfPhase,EndOfPhase,StartOfGame,EndOfGame,StartOfMatch,EndOfMatch,StartOfSession,EndOfSession,Sites,Moves,All,Angled,Adjacent,Axial,Orthogonal,Diagonal,Off,SameLayer,Upward,Downward,Rotational,N,E,S,W,NE,SE,NW,SW,NNW,WNW,WSW,SSW,SSE,ESE,ENE,NNE,CW,CCW,In,Out,U,UN,UNE,UE,USE,US,USW,UW,UNW,D,DN,DNE,DE,DSE,DS,DSW,DW,DNW,N,NNE,NE,ENE,E,ESE,SE,SSE,S,SSW,SW,WSW,W,WNW,NW,NNW,N,NNE,NE,E,SSE,SE,S,SSW,SW,W,NW,NNW,WNW,ENE,ESE,WSW,CW,Out,CCW,In,UNW,UNE,USE,USW,DNW,DNE,DSE,DSW,U,UN,UW,UE,US,D,DN,DW,DE,DS,Forward,Backward,Rightward,Leftward,Forwards,Backwards,Rightwards,Leftwards,FL,FLL,FLLL,BL,BLL,BLLL,FR,FRR,FRRR,BR,BRR,BRRR,SameDirection,OppositeDirection,Out,CW,In,CCW,D,DN,DNE,DE,DSE,DS,DSW,DW,DNW,U,UN,UNE,UE,USE,US,USW,UW,UNW,Checkered,Colour,Shape,StyleThickness,Style,Board,Animation,HandScale,Curves,MaskedColour,ColourFromState,Colour,Families,Background,Foreground,Rename,ExtendName,AddStateToName,Reflect,ByValue,Scale,Style,Colour,Colour,Pits,PlayerHoles,RegionOwner,Cost,Check,State,Value,Piece,Edges,Score,Shape,AsHoles,Sites,Cell,Symbol,InnerEdges,OuterEdges,Phase0,Phase1,Phase2,Phase3,Symbols,Vertices,White,Black,Grey,LightGrey,VeryLightGrey,DarkGrey,VeryDarkGrey,Dark,Red,Green,Blue,Yellow,Pink,Cyan,Brown,DarkBrown,Purple,Turquoise,Orange,DarkOrange,LightRed,DarkRed,LightGreen,DarkGreen,LightBlue,VeryLightBlue,DarkBlue,IceBlue,Gold,Silver,Bronze,GunMetal,HumanLight,HumanDark,Cream,DeepPurple,PinkFloyd,BlackSabbath,KingCrimson,MoodyBlues,TangerineDream,Piece,Tile,Card,Die,Domino,LargePiece,ExtendedShogi,Board,Hand,Deck,Dice,Boardless,ConnectiveGoal,Mancala,PenAndPaper,Pyramidal,Spiral,Isometric,Puzzle,GraphPuzzle,LineSegment,PuzzlePenAndPaper,RegionPuzzle,Agon,Backgammon,Chess,ChineseCheckers,Connect4,Goose,Go,Graph,HoundsAndJackals,Janggi,Lasca,Pachisi,Ploy,Scripta,Shogi,SnakesAndLadders,Surakarta,Tafl,Xiangqi,UltimateTicTacToe,Futoshiki,Hashi,Kakuro,Sudoku,BasicController,PyramidalController,All,Inner,Outer,Interlayer,Thin,Thick,ThinDotted,ThickDotted,ThinDashed,ThickDashed,Hidden,Default,Ground,Reverse,Fan,None,Backgammon,Ring,None,Corner,Middle,Always,Never,AtEnd},
  ndkeywordstyle=\color{dviolet}\bfseries,
  identifierstyle=\color{black},
  sensitive=true,   % need case-sensitivity for different keywords
  comment=[l]{//},
  commentstyle=\color{dred}\ttfamily,
  stringstyle=\color{dgreen}\ttfamily,
  morestring=[b]',
  morestring=[b]",
  escapechar=@,
  showstringspaces=false,
  xleftmargin=1pt,xrightmargin=1pt,
  breaklines=true,basicstyle=\ttfamily\small,backgroundcolor=\color{colorex},inputencoding=utf8/latin9,texcl
}

\newcommand{\core}{ 
  \medskip \begin{tcolorbox}[
    enhanced,breakable,
    boxsep=0pt,top=0pt,bottom=0pt,left=7mm,right=1mm,
    toprule=0.1mm,leftrule=0.1mm,rightrule=0.25mm,bottomrule=0.25mm,shadow={0.2mm}{-0.2mm}{0mm}{dgray},
    overlay unbroken and first={\node (logo) at ([xshift=6mm,yshift=-5mm]frame.north west) {}; %\draw[black,line width=1.5pt] (logo) -- ([xshift=4mm,yshift=1.5mm]frame.south west);  
    },
    colframe=dgray,titlerule=-0.2mm,toptitle=3mm,coltitle=black,fonttitle=\bfseries,
    lines before break=6, pad at break*=10pt
}
\newcounter{cntEx}
\newcounter{cntNaturalLanguage}

\ifx\bw\undefined
  \lstnewenvironment{ludii}[1][]{\lstset{language=ludii,#1}}{}
  \lstnewenvironment{syntax}{\lstset{}}{}
  \newenvironment{boxex}
    {\stepcounter{cntEx} \core ,colback=colorex,title style={color=colorex},title=Ludii Game Description Example %\thecntEx
    ]}
    {\end{tcolorbox}} %{\vspace{-0.1cm}\end{tcolorbox} ~ \vspace{-0.2cm}}
    \newenvironment{boxnaturallanguage}
    {\stepcounter{cntNaturalLanguage} \core ,colback=colorex,title style={color=colorex},title=Natural Language Description Example %\thecntNaturalLanguage
    ]}
    {\end{tcolorbox}} %{\vspace{-0.1cm}\end{tcolorbox} ~ \vspace{-0.2cm}}

\definecolor{v2lgray}{gray}{0.85}
\definecolor{dgray}{rgb}{0.4,0.4,0.4}
\definecolor{dblue}{RGB}{0,0,99}
\definecolor{dred}{RGB}{150,6,54}
\definecolor{dgreen}{RGB}{47,135,7}
\definecolor{dviolet}{RGB}{102,0,153}
\definecolor{mblue}{RGB}{0,0,180}
\definecolor{colorex}{HTML}{DFEFFF}  %{FFE3BE}
\definecolor{grey1}{rgb}{0.9,0.9,0.9}

\begin{document}
\title{A Research Agenda for Usability and Generalisation in Reinforcement Learning}
\titlerunning{A Research Agenda for Usability and Generalisation in RL}
% If the paper title is too long for the running head, you can set
% an abbreviated paper title here
%
\author{Dennis J.N.J. Soemers\inst{1} \and
Spyridon Samothrakis\inst{2} \and \\
Kurt Driessens\inst{1} \and Mark H.M. Winands\inst{1}}
\authorrunning{D.J.N.J. Soemers et al.}
% First names are abbreviated in the running head.
% If there are more than two authors, 'et al.' is used.
%
\institute{Department of Advanced Computing Sciences, Maastricht University,\\ Maastricht, the Netherlands \\
\email{\{dennis.soemers, kurt.driessens, m.winands\}@maastrichtuniversity.nl}\\ 
\and
Institute for Analytics and Data Science, University of Essex,\\ Colchester, United Kingdom\\
\email{ssamot@essex.ac.uk}}
\maketitle              % typeset the header of the contribution
\begin{abstract}
It is common practice in reinforcement learning (RL) research to train and deploy agents in bespoke simulators, typically implemented by engineers directly in general-purpose programming languages or hardware acceleration frameworks such as CUDA or JAX. This means that programming and engineering expertise is not only required to develop RL algorithms, but is also required to use already developed algorithms for novel problems. The latter poses a problem in terms of the usability of RL, in particular for private individuals and small organisations without substantial engineering expertise. We also perceive this as a challenge for effective generalisation in RL, in the sense that is no standard, shared formalism in which different problems are represented. As we typically have no consistent representation through which to provide information about any novel problem to an agent, our agents also cannot instantly or rapidly generalise to novel problems. In this position paper, we advocate for a research agenda centred around the use of user-friendly description languages for describing problems, such that (i) users with little to no engineering expertise can formally describe the problems they would like to be tackled by RL algorithms, and (ii) algorithms can leverage problem descriptions to effectively generalise among all problems describable in the language of choice.
%The majority of current reinforcement learning (RL) research involves training and deploying agents in environments that are implemented by engineers in general-purpose programming languages, possibly in combination with more advanced frameworks such as CUDA or JAX. In contrast to many recent advances in generative artificial intelligence, which are often made easily accessible through convenient user interfaces, this makes the application of RL to novel problems of interest inaccessible to small organisations or private individuals without such expertise. This position paper argues that, to enable more widespread adoption of RL, it is important for the research community to shift more focus towards methodologies where environments are described in user-friendly domain-specific languages, or, ultimately, natural languages. Aside from improving the usability of RL, such language-based environment descriptions may also provide valuable context and boost the ability of trained agents to generalise to unseen environments within the set of all environments that can be described in any language of choice.
%
%The abstract should briefly summarize the contents of the paper in 150--250 words.

\end{abstract}

\section{Introduction}

Researchers in the field of reinforcement learning (RL) have largely converged on a small number of common APIs for the development of benchmark domains that are used to evaluate the performance of RL algorithms. New environments are customarily written in general-purpose programming languages such as \texttt{C++} or \texttt{Python}, and implement a Gym-based \cite{Brockman_2016_OpenAIGym} API for learning algorithms to interface with environments. There are variants on this approach, such as PettingZoo \cite{Terry_2021_PettingZoo} for multi-agent RL, but the general workflow remains similar.

At a high level, we may distinguish roughly two categories of RL research. On the one hand, there is research where the focus is on the development of new and improved (modifications of) training algorithms, typically not focused on any specific task. There may be a focus on certain categories of tasks (single-agent RL, multi-agent RL, RL for domains with vision-based inputs, RL for partially observable environments, RL for continuous action spaces, and so on), but existing and established frameworks with a suite of applicable domains are typically used for empirical evaluations. Researchers typically aim to demonstrate a high level of generality, by showing that an algorithm can effectively learn on a large collection of different environments within such a suite, as opposed to only a single environment. These environments are often games or other simulations, with arguably relatively limited direct real-world impact outside of their use as benchmarks for RL research. Examples of such suites of environments include the Arcade Learning Environment \cite{Bellemare_2013_ALE,Machado_2018_Revisiting}, ProcGen \cite{Cobbe_2020_ProcGen}, SMACv2 \cite{Ellis_2023_SMACv2}, and the DeepMind Control Suite \cite{Tassa_2018_DeepMind}. On the other hand, there is research in which a single, concrete, high-impact ``real-world'' task is selected, and RL is used to improve performance on that one task. Substantial engineering effort is often dedicated towards implementing and optimising such a task for the purpose of this research. This engineering effort often requires highly specialised knowledge of, for example, programming for GPUs or other hardware accelerators, or of the inner workings of deep learning (DL) and RL algorithms, such that the task and its state and action representations can be implemented in such a way that the existing DL and RL techniques can be applied to their fullest potential. Examples include research focused on applications such as resource balancing for logistics \cite{Li_2019_Cooperative}, flight control for stratospheric balloons \cite{Bellemare_2020_Autonomous}, compiler optimisations \cite{Cummins_2021_CompilerGym,Trofin_2021_MLGO}, automated chip floorplanning \cite{Mirhoseini_2021_Graph}, power grid management \cite{Yoon_2021_PowerGrid}, alignment of large language models with human preferences \cite{Ouyang_2022_Training,Kaufmann_2025_SurveyRLHF}, and magnetic tokamak controllers \cite{Degrave_2022_Magnetic}.

\begin{figure}
\centering
\begin{subfigure}{.31\textwidth}
\centering
\resizebox{\linewidth}{!}{
\begin{tikzpicture}
\node[draw,shape=rectangle,rounded corners,align=center,orange!75!black,fill=orange!15!white,dashed] (Agent) at (0, 0) {\textbf{Agent}};
\node[draw,shape=rectangle,dashed,rounded corners,right=1.5cm of Agent,text width=20mm,align=center,orange!75!black,fill=orange!15!white,node distance=1.5cm] (Simulator) {\textbf{Simulator}};

% ONLY HERE FOR PADDING
\node[shape=rectangle,rounded corners,align=center,orange!75!black,dashed,above=0.5cm of Simulator] (Compiler) {};
% ONLY HERE FOR PADDING
\node[shape=document,align=center,minimum width=12mm,minimum height=20mm,above=0.5cm of Compiler,green!75!black] (EnvDescription) {};

\path[draw,-{triangle 90},out=270,in=270] (Simulator.south) to node[midway,below,align=center] {\textit{State Observations,}\\\textit{Rewards}} (Agent.south);
\path[draw,-{triangle 90},out=0,in=180] (Agent.east) to node[midway,above,align=center] {\textit{Actions}} (Simulator.west);
\end{tikzpicture}
}
\caption{A simulator of the environment is implemented in a general-purpose programming language. This is the most common setting in current research.}
\label{Fig:ProgrammaticEnvironment}
\end{subfigure}
\hfill
\begin{subfigure}{.31\textwidth}
\centering
\resizebox{\linewidth}{!}{
\begin{tikzpicture}
\node[draw,shape=rectangle,rounded corners,align=center,orange!75!black,fill=orange!15!white,dashed] (Agent) at (0, 0) {\textbf{Agent}};

\node[draw,shape=rectangle,rounded corners,right=1.5cm of Agent,text width=20mm,align=center,blue!75!black,fill=blue!15!white,node distance=1.5cm] (Simulator) {\textbf{Simulator}};
\node[draw,shape=rectangle,rounded corners,align=center,orange!75!black,fill=orange!15!white,dashed,above=0.5cm of Simulator] (Compiler) {\textbf{Compiler}};
\node[shape=document,draw,align=center,minimum width=12mm,minimum height=20mm,above=0.5cm of Compiler,green!75!black,fill=green!15!white] (EnvDescription) {\textbf{User-friendly}\\\textbf{Environment}\\\textbf{Description}};

\draw[pink,ultra thick,dotted,rounded corners] ($(Compiler.north west)+(-0.38,2.65)$) rectangle ($(Simulator.north east)+(0.1,0.35)$);

\path[draw,-{triangle 90}] (EnvDescription.south) to (Compiler.north);
\path[draw,-{triangle 90}] (Compiler.south) to (Simulator.north);

\path[draw,-{triangle 90},out=270,in=270] (Simulator.south) to node[midway,below,align=center] {\textit{State Observations,}\\\textit{Rewards}} (Agent.south);
\path[draw,-{triangle 90},out=0,in=180] (Agent.east) to node[midway,above,align=center] {\textit{Actions}} (Simulator.west);
\path[draw,-{triangle 90},out=180,in=90,dashed] (EnvDescription.west) to node[midway,left,align=center] {\textit{Context}} (Agent.north);
\end{tikzpicture}
}
\caption{The environment is described in a user-friendly language, and compiled into a simulator. The description may serve as additional input to the agent.}
\label{Fig:DSLEnvironment}
\end{subfigure}
\hfill
\begin{subfigure}{.35\textwidth}
\centering
\resizebox{\linewidth}{!}{
\begin{tikzpicture}
\node[draw,shape=rectangle,rounded corners,right=1.5cm of Agent,text width=65mm,align=center,blue!75!black,fill=blue!15!white,node distance=1.5cm] (Simulator) {\textbf{Simulator}};

\node[draw,shape=rectangle,rounded corners,align=center,orange!75!black,fill=orange!15!white,dashed,above=1cm of Simulator.west,anchor=west] (Compiler1) {\textbf{Compiler}};
\node[shape=document,draw,align=center,minimum width=12mm,minimum height=12mm,above=0.5cm of Compiler1,green!75!black,fill=green!15!white] (EnvDescription1) {\textbf{DSL}\\\textbf{Description}};
\draw[pink,ultra thick,dotted,rounded corners] ($(Compiler1.north west)+(-0.3,1.85)$) rectangle ($(Compiler1.south east)+(0.25,-0.15)$);

\node[draw,shape=rectangle,rounded corners,align=center,orange!75!black,fill=orange!15!white,dashed,above=0.5cm of Simulator] (Compiler2) {\textbf{Compiler}};
\node[shape=document,draw,align=center,minimum width=12mm,minimum height=12mm,above=0.5cm of Compiler2,blue!75!black,fill=blue!15!white] (IntermediateDSLDescription) {\textbf{DSL}\\\textbf{Description}};
\node[draw,shape=rectangle,rounded corners,align=center,orange!75!black,fill=orange!15!white,dashed,above=0.5cm of IntermediateDSLDescription] (LLM2) {\textbf{Language}\\\textbf{Model}};
\node[shape=document,draw,align=center,minimum width=12mm,minimum height=12mm,above=0.5cm of LLM2,green!75!black,fill=green!15!white] (EnvDescription2) {\textbf{Natural}\\ \textbf{Language}\\\textbf{Description}};
\draw[pink,ultra thick,dotted,rounded corners] ($(LLM2.north west)+(-0.3,2)$) rectangle ($(Compiler2.south east)+(0.25,-0.15)$);

\node[draw,shape=rectangle,rounded corners,align=center,orange!75!black,fill=orange!15!white,dashed,above=4cm of Simulator.east,anchor=east] (LLM3) {\textbf{Language}\\\textbf{Model}};
\node[shape=document,draw,align=center,minimum width=12mm,minimum height=12mm,above=0.5cm of LLM3,green!75!black,fill=green!15!white] (EnvDescription3) {\textbf{Natural}\\ \textbf{Language}\\\textbf{Description}};
\draw[pink,ultra thick,dotted,rounded corners] ($(LLM3.north west)+(-0.3,2)$) rectangle ($(LLM3.south east)+(0.25,-0.15)$);

\draw[-{triangle 90}] (EnvDescription1.south) -- (EnvDescription1.south |- Compiler1.north);
\draw[-{triangle 90}] (Compiler1.south) -- (Compiler1.south |- Simulator.north);

\draw[-{triangle 90}] (EnvDescription2.south) -- (EnvDescription2.south |- LLM2.north);
\draw[-{triangle 90}] (LLM2.south) -- (LLM2.south |- IntermediateDSLDescription.north);
\draw[-{triangle 90}] (IntermediateDSLDescription.south) -- (IntermediateDSLDescription.south |- Compiler2.north);
\draw[-{triangle 90}] (Compiler2.south) -- (Compiler2.south |- Simulator.north);

\draw[-{triangle 90}] (EnvDescription3.south) -- (EnvDescription3.south |- LLM3.north);
\draw[-{triangle 90}] (LLM3.south) -- (LLM3.south |- Simulator.north);
\end{tikzpicture}
}
\caption{A DSL description can be converted into a simulator by a compiler. A natural language description can first be translated into an intermediate DSL description by a language model, or can be translated directly with a language model serving as compiler.}
\label{Fig:CompilerTypes}
\end{subfigure}
\caption{Orange boxes with dashed lines represent components that require substantial engineering or RL expertise. The green components can be provided by users with little to no engineering expertise. \textbf{(a)} A depiction of the customary setting in RL research. \textbf{(b)} The approach for which we posit that increased research attention is warranted (\refsection{Sec:DescriptionLanguagesEnvironments}). \textbf{(c)} User-friendly environment descriptions may be written in a DSL, or in a natural language, where the latter approach may or may not generate an intermediate DSL description. Image source: \cite{Soemers_2025_EnvironmentDescriptions}.}
\label{Fig:EnvironmentDefinitionStyles}
\end{figure}

While discussions on experimental methodologies and statistical analyses of empirical evaluations in RL have been on the rise within the research community \cite{Henderson_2018_DeepRLThatMatters,Agarwal_2021_Deep,Jordan_2022_Scientific,Patterson_2023_Empirical,Jordan_2024_Position}, as well as discussions of how to select which environment(s) to use in experiments \cite{Patterson_2023_Empirical,Voelcker_2024_Hop}, we see little to no discussion on \emph{how} or \emph{by whom} tasks (or environments) are described or implemented in the first place. In a recent position paper \cite{Soemers_2025_EnvironmentDescriptions}, we argued that the standard assumption that environments can be implemented (and heavily optimised) in general-purpose programming languages, by engineers familiar with machine learning, (i) poses a challenge to widespread adoption of RL for real-world use cases, and (ii) also leads the research community at large to miss out on interesting research directions with respect to generalisation and transfer in RL. While it may be acceptable to invest substantial engineering resources for the implementation of environments for large-scale projects with high potential impact, it impedes the application of RL by smaller organisations or private individuals. We posit that more widespread applications of RL will be greatly aided if the latter groups can express their tasks in user-friendly domain-specific languages (DSLs) \cite{Mernik_2005_DSLs,Aram_2015_Multilayered}, or even in natural language. 
%There are many possible definitions and interpretations of the term ``user-friendly'' \cite{Stevens_1983_UserFriendly}, but as a working definition, we will say that \emph{a language is user-friendly if it is easy to use for users who may be experts in their application domain of interest, but may not have any RL, AI, or programming expertise, and is designed with their needs in mind}.

Once we adopt a methodology in which environments are represented in explicit forms that can be provided as inputs to an agent (e.g., DSL or natural language snippets), we can also explore new forms of generalisation or transfer in RL, where effective generalisation or zero-shot transfer to unseen environments may become feasible given sufficient understanding of the task descriptions. \reffigure{Fig:ProgrammaticEnvironment} depicts the setting where the environment is implemented directly in a general-purpose programming language, and \reffigure{Fig:DSLEnvironment} depicts the proposed settings, with \reffigure{Fig:CompilerTypes} providing three examples of how the translation from a user-friendly environment description to a simulator may work. 
%\refappendix{Appendix:ExampleDescriptions} provides example descriptions in a DSL, as well as natural language, for the game of Tic-Tac-Toe as an environment.
For tasks that take place in the physical world, such as non-simulated robotics tasks, a description of the reward function can suffice, as hardware and the real world with its laws of physics already dictate aspects such as the action space and transition dynamics. However, even in these cases, the ability to automatically generate a sufficiently accurate simulator from user-friendly descriptions would, in combination with sim-to-real transfer \cite{Zhao_2020_SimToReal}, still be highly beneficial \cite{Yang_2024_Learning}.

This paper is an extension of a previous position paper \cite{Soemers_2025_EnvironmentDescriptions}. The primary extensions consist of (i) a discussion of barriers to widespread adoption of RL other than the one that is our main focus (\refsection{Sec:OtherBarriers}), and (ii) a detailed description of the research agenda we envision following from our position (\refsection{Sec:ProposedResearchAgenda}). Furthermore, other sections throughout the paper have been updated to provide additional context, examples, and to reflect and relate our work to other recent developments in the field, such as the Ludax framework \cite{Todd_2025_Ludax} and other related position papers \cite{Rolnick_2024_ApplicationDriven,BlaliHamelin_2025_AGINorthStar,Hartman_2025_ResponsibleApplicationDriven}.

\section{Background} \label{Sec:Background}

This section provides background information on (partially observable) Markov decision processes \cite{Howard_1960_Dynamic} and Markov games \cite{vanderWal_1981_Stochastic} (\refsubsection{Subsec:MDPsAndMarkovGames})---two frequently used formalisations of sequential decision-making problems, which we argue users should be able to describe in more accessible languages than programming languages---and deep reinforcement learning (\refsubsection{Subsec:DeepRL}).

\subsection{Markov Decision Processes \& Markov Games} \label{Subsec:MDPsAndMarkovGames}

A Markov decision process (MDP) $\mathcal{M} = \langle \mathcal{S}, \mathcal{A}, P, \rho \rangle$ is a formal description of a single-agent sequential decision-making problem, in which $\mathcal{S}$ denotes the state space, $\mathcal{A}$ the action space, $P : \mathcal{S} \times \mathcal{A} \times \mathcal{S} \mapsto \left[ 0, 1 \right]$ a function that defines transition probabilities, and $\rho : \mathcal{S} \times \mathcal{A} \times \mathcal{S} \times \mathbb{R}$ a reward function. More precisely, $0 \leq P(s' \mid s, a) \leq 1$ gives the probability of observing a transition from a current state $s \in \mathcal{S}$ to a successor state $s' \in \mathcal{S}$ after executing an action $a \in \mathcal{A}$, and $\rho(r \mid s, a, s')$ denotes the probability of observing a real-valued reward $r \in \mathbb{R}$ after the same event. In some MDPs, it may be the case that only certain subsets of the full action space $\mathcal{A}$ are legal in certain states. 

The behaviour of an agent may be described as a policy $\pi : \mathcal{S} \times \mathcal{A} \mapsto \left[ 0, 1 \right]$, such that $0 \leq \pi(a \mid s) \leq 1$ denotes the probability with which the policy selects an action $a \in \mathcal{A}$ whenever the current state is $s \in \mathcal{S}$. At discrete time steps $t = 0, 1, \dots$, the agent observes the current state $S_t \in \mathcal{S}$, samples an action $A_t \sim \pi(\cdot \mid S_t)$ from its policy $\pi$, transitions into a successor state $S_{t+1} \sim P(\cdot \mid S_t, A_t)$, and receives a reward $R_t \sim \rho(S_t, A_t, S_{t+1})$. The most common objective is to select actions such that the returns $G_0$ are maximised, where $G_t = \sum_{k=0}^{\infty} \gamma^k R_{t+k+1}$ denotes sum of discounted rewards collected from time $t$ onwards. Temporally distant rewards are discounted, relative to short-term rewards, by the discount factor $0 < \gamma \leq 1$.

The \textit{state value function} $V^{\pi} : \mathcal{S} \mapsto \mathbb{R}$ gives the expected returns when action according to a policy $\pi$ from any input state $s$ onwards: $V^{\pi}(s) = \mathbb{E}_{\pi} \left[ G_t \mid S_t = s \right]$. Similarly, the \textit{state-action value function} $Q^{\pi} : \mathcal{S} \times \mathcal{A} \mapsto \mathbb{R}$ gives the expected returns of executing an input action $a$ in an input state $s$, and sampling any subsequent actions from $\pi$: $Q^{\pi}(s, a) = \mathbb{E}_{\pi} \left[ G_t \mid S_t = s, A_t = a \right]$. 

\emph{Partially observable} Markov decision processes (POMDPs) additionally feature an observation space $\mathcal{O}$, and a mapping $\phi : \mathcal{S} \mapsto \mathcal{O}$ from states to observations. In a POMDP, the agent cannot necessarily observe the current state $S_t$, but only an observation $O_t = \phi(S_t)$, where it is possible for multiple different states to map to the same observation.

The concept of Markov games may be viewed as a generalisation of (PO)MDPs for multi-agent settings \cite{Littman_1994_Markov}. In a Markov game, each of $k \geq 1$ players (or agents) has its own action space from which to select actions, and its own reward function. Transition probabilities between pairs of states are defined over the joint action space of all players. This paper uses the term \textit{environment} to refer to problems that may be tackled by RL, regardless of whether they may be (PO)MDPs or Markov games.

\subsection{Deep Reinforcement Learning} \label{Subsec:DeepRL}

One of the main goals in reinforcement learning (RL) \cite{Sutton_2018_RL} research is the development of algorithms that can automatically learn strong policies from experience gathered within a (PO)MDP or Markov game. 
%Two important categories of RL methods are \textit{value-based} methods and \textit{policy-based methods}. Value-based methods aim to learn approximations of state or state-action value functions, and implement policies that select actions according to value estimates from these approximations. Policy-based methods directly optimise a parameterised policy, typically building on the policy gradient theorem \cite{Sutton_2000_PolicyGradient}. 
\textit{Tabular} RL methods learn distinct state-action values or probabilities for every individual state-action pair in $\mathcal{S} \times \mathcal{A}$. This is only feasible in relatively small problems, and often wasteful in terms of data efficiency due to the lack of generalisation between related state-action pairs. RL methods with \textit{function approximation} address these issues, by training function approximators that use features of states and/or actions, rather than enumerating the complete space. State-action pairs that are closely related to each other may be expected to have similar features, allowing for improved data efficiency through generalisation. One of the most popular and successful forms of RL with function approximation is deep RL, in which deep neural networks (DNNs) \cite{LeCun_2015_DeepLearning} are used as function approximators.

\section{Description Languages for Environments} \label{Sec:DescriptionLanguagesEnvironments}

\refsubsection{Subsec:StandardPractice} describes the established practice where RL research is performed using environments that have been implemented, conforming to a standardised API such as the Gym API \cite{Brockman_2016_OpenAIGym}, in general-purpose programming languages or more specialised frameworks for hardware accelerators such as GPUs. It identifies potential issues that may arise when the entire research community focuses solely on such environments. As an initial step towards more user-friendly descriptions, \refsubsection{Subsec:DSLMDPs} discusses the use of DSLs for describing environments for use in RL research. \refsubsection{Subsec:NaturalLanguageMDPs} explores the possibility of using natural languages to define environments---arguably one of the most user-friendly language classes. Finally, \refsubsection{Subsec:CallBenchmarks} presents the central position of this paper: a call for more (research attention for) benchmarks in which environments are described in DSLs or natural language.

\subsection{Defining Environments in Programming Languages} \label{Subsec:StandardPractice}

Outside of cases where RL is applied directly in the physical world, such as some work in robotics \cite{Luo_2023_SERL}, it is customary to implement the environments used for RL research in programming languages such as \texttt{C++} or \texttt{Python}. There is also a recent trend of using more advanced toolkits or libraries, such as CUDA or JAX \cite{Bradbury_2018_JAX}, to enable the environments themselves---and not just DNN forward and backward passes---to make efficient use of hardware accelerators \cite{Dalton_2020_Accelerating,Freeman_2021_Brax,Makoviychuk_2021_Isaac,Bettini_2022_VMAS,Lange_2022_Gymnax,Frey_2023_JAXLOB,Gulino_2023_Waymax,Koyamada_2023_Pgx,Rutherford_2023_JaxMARL,Bonnet_2024_Jumanji,Ponse_2025_Chargax}. The latter approach can provide dramatic speed increases, but also imposes additional constraints on programming style and requires more specialised and advanced engineering skills.

Most developers of RL environments have converged to the API popularised by OpenAI Gym \cite{Brockman_2016_OpenAIGym}. This API requires developers to implement:
\begin{itemize}
    \item A definition of the \textit{observation space}. For any state that an agent may ever reach in an environment, it will receive an observation from this space as input. The observation space may, for example, be a discrete set of integers, each of which uniquely identifies one element of the state space $\mathcal{S}$, or it may be subset of $\mathbb{R}^d$ for some dimensionality $d$, such that every state is described by a $d$-dimensional real-valued feature vector.
    \item A definition of the \textit{action space} $\mathcal{A}$. It is typically assumed that agents must select any one element from this space as their action at each time step.
    \item A \textit{reset} function, which resets the environment to an initial state.
    \item A \textit{step} function, which takes an action from $\mathcal{A}$ as input, transitions from a current state $s \in \mathcal{S}$ to a successor state $s' \in \mathcal{S}$, and returns a real-valued reward $r$ and an observation of $s'$ (alongside several auxiliary variables). In this function, programmers essentially implement the transition and reward models $P$ and $\rho$. This is typically done implicitly: the function usually implements a procedural algorithm that generates $s'$ and $r$ in a manner that ends up being consistent with the probabilities defined by $P$ and $\rho$, without explicitly defining and sampling from the full distributions.
\end{itemize}

\subsection{Domain-Specific Languages for Environments} \label{Subsec:DSLMDPs}

A potential alternative to the standard practice of describing environments in general-purpose (or more advanced and complex hardware acceleration frameworks) is to use DSLs to describe sets of environments. This approach still requires significant engineering effort to develop a compiler that can translate valid descriptions from the DSL to a runnable simulator with an API for (learning) agents. However, once this compiler has been built, users with little to no programming experience may---depending on the complexity and user-friendliness of the DSL in question---use it to describe new environments that fit within the overarching domain supported by the DSL. Note that the requirement for a compiler to be implemented suggests that it would only ever be reasonable to use a DSL for sets of multiple environments, but never for a single environment. If there is only a single environment of interest, it would likely be easier to directly implement that environment itself in a programming language.

Numerous examples of DSLs for describing sequential decision-making problems already exist, though their adoption as benchmarks within the RL community is restricted compared to benchmarks such as the Arcade Learning Environment \cite{Bellemare_2013_ALE,Machado_2018_Revisiting} or the DeepMind Control Suite \cite{Tassa_2018_DeepMind}, which are not based on DSLs. Examples include the Planning Domain Definition Language (PDDL) \cite{McDermott_1998_PDDL} for planning problems, and the Stanford Game Description Language \cite{Love_2008_GDL,Genesereth_2014_GGP}, Ludi \cite{Browne_2009_PhD}, Toss \cite{Kaiser_2011_FirstOrder}, the Video Game Description Language \cite{Schaul_2013_VGDL,Schaul_2014_Extensible}, Regular Boardgames \cite{Kowalski_2019_Regular}, Ludii \cite{Piette_2020_Ludii}, Stratega \cite{Perez_2020_Stratega}, Griddly \cite{Bamford_2020_Griddly}, MiniHack \cite{Samvelyan_2021_MiniHack}, and Ludax \cite{Todd_2025_Ludax} for various ranges of games. PDDLGym \cite{Silver_2020_PDDLGym} provides Gym environment wrappers around PDDL problems. 

As an example, we provide a description of the game of Tic-Tac-Toe in the DSL of Ludii below. Note that some rules---such as players moving in turns, or the game ending in a draw when neither player can move---are not explicitly stated, because these are default settings and can therefore be omitted in this language. Games that deviate from these defaults can be described by explicitly including rules where appropriate.
\begin{boxex}
\begin{ludii}
(game "Tic-Tac-Toe"  
    (players 2)  
    (equipment { 
        (board (square 3)) 
        (piece "Disc" P1) 
        (piece "Cross" P2) 
    })  
    (rules 
        (play (move Add (to (sites Empty))))
        (end (if (is Line 3) (result Mover Win)))
    )
)
\end{ludii} 
\end{boxex}

\subsection{Describing Environments in Natural Language} \label{Subsec:NaturalLanguageMDPs}

While DSLs may already be considered a more user-friendly alternative to general-purpose programming languages \cite{Mernik_2005_DSLs,Aram_2015_Multilayered} for describing environments, natural language would be even more accessible to a wider userbase. Here, we provide an example of what a natural language description of the game of Tic-Tac-Toe might look like:

\begin{boxnaturallanguage}
\vspace{6pt}
The game of Tic-Tac-Toe is played by two players on a grid of $3$$\times$$3$ square cells. Players take turns drawing their symbol---a circle for the first player, and a cross for the second---in an empty cell of their choice. The game ends in a win for a player if that player completes an orthogonal or diagonal line of three instances of their symbol. The game ends in a draw if the board is filled up with neither player achieving their win condition.
\vspace{6pt}
\end{boxnaturallanguage}

Although the state of the art of large language models (LLMs) is highly impressive \cite{Zhao_2023_survey}, there are still concerns surrounding reliability and correctness \cite{Leivada_2024_Sentence}. In particular, we envision that ambiguities typically present in natural languages, as well as the tendency for humans to underspecify task descriptions (e.g., rules of games), will present substantial challenges that require further research. Recently, \cite{Afshar_2024_DeLF} demonstrated promising initial results for an LLM automatically generating executable environment code based on natural language descriptions, but it still requires an expert human in the loop who is able to interpret the generated code and provide feedback on potential mistakes. In the short term, it may be more realistically feasible to use a combination of natural language and DSLs, where an LLM first translates a natural language description to a DSL-based description \cite{Desai_2016_Program,Oswald_2024_Large,Zuo_2024_Planetarium}, and a user can inspect the generated description and make corrections if necessary before it is compiled into a simulator. In the long term, if LLMs can be made sufficiently reliable, natural languages would likely be the most accessible modality for describing environments.

\subsection{Research Focus on Description Languages for Environments} \label{Subsec:CallBenchmarks}

Before formally stating the central position of this paper, we make two core assumptions relating to the user-friendliness of DSLs and natural languages (\refassumption{Assumption:MoreUserFriendly}), and the desirability of this user-friendliness (\refassumption{Assumption:UserFriendlinessGood}).

\begin{assumption} \label{Assumption:MoreUserFriendly}
    Defining environments, such as (PO)MDPs or Markov games, in domain-specific languages (DSLs) or natural languages can be more user-friendly than defining them in general-purpose programming languages.
\end{assumption}

Increasing user-friendliness and lowering barriers to entry is a well-established potential motivation for the use of DSLs \cite{Mernik_2005_DSLs,Aram_2015_Multilayered}. Note that there may also be other reasons for using DSLs, and there can be DSLs that do not substantially lower barriers to entry: this depends on the design of the DSL in question. For example, the logic-based Stanford Game Description Language \cite{Love_2008_GDL,Genesereth_2014_GGP} arguably still requires substantial technical expertise and writing games in it may be considered error-prone due to the large file size required for many interesting games. In contrast, allowing for clear and succinct descriptions that are easy to read and write was an explicit design goal for Ludii's description language \cite{Piette_2020_Ludii}. Likely in no small part due to the language's level of accessibility, Ludii has amassed a library of over 1400 distinct official game descriptions,\footnote{\url{https://ludii.games/library.php}} including third-party contributions from game designers with little or no programming experience.\footnote{\url{https://ludii.games/forum/forumdisplay.php?fid=23}} 

In the case of natural languages, if any concerns around ambiguities and underspecification of environments can be adequately addressed, we see little reason to doubt that many users would indeed find them more accessible than programming languages. If procedures translating natural language descriptions directly into executable simulations cannot be made sufficiently reliable, a potential solution may be to use DSLs as an intermediate step. Users could first describe their tasks in natural languages, and they would ideally only have to verify or fix small issues in automatically generated descriptions in a user-friendly DSL afterwards.

\begin{assumption} \label{Assumption:UserFriendlinessGood}
    Enabling environments to be defined in more user-friendly ways is desirable.
\end{assumption}

First, we will acknowledge that lowering the barrier to entry for defining environments is not necessarily \textit{always} of importance. For example, when RL is applied to an individual, specific domain with a high degree of scientific, societal, economic or other form of impact, it will often be worth investing substantial engineering effort into the environment definition. However, running such projects tends to be restricted to groups with direct access to RL experts. 

A survey among AI engineers, AI designers, and RL engineers from AAA video game studios, independent developers, and industrial research labs---most of which do have direct access to a substantial amount of engineering expertise---revealed, among other concerns, an overreliance on engineering support, and difficulties in designing tasks for RL agents, as challenges for the adoption of RL and other AI techniques in video game development \cite{Jacob_2020_Unwieldy}. While not focused on RL (but, rather, AI in general) or environment descriptions, a recent study by \cite{Simkute_2024_Adoption} reveals a substantial disconnect between the technical know-how that AI designers expect users will have, and what they actually tend to have, as a core barrier to adoption of AI in practice. These studies point to the relevance of improving the user-friendliness of any aspect of the RL (or any AI) pipeline.

If we wish to democratise the use of AI \cite{Seger_2023_Democratising} to the extent that users with little expertise in RL---or even programming---can apply it to their problems of interest, enabling environments to be defined in more user-friendly ways would be a requirement. Outside of RL, in the landscape of generative artificial intelligence (AI), substantial value is generated not necessarily just by the models themselves, but also by the release of user-friendly tools and interfaces to access the trained models. Famous examples include OpenAI's ChatGPT \cite{OpenAI_2022_ChatGPT} and DALL$\cdot$E 2 \cite{Ramesh_2022_Dalle2}, Stability AI's models and interfaces for audio \cite{Stability_2023_Audio}, image \cite{Podell_2023_SDXL}, and video \cite{Blattmann_2023_StableVideo} generation, Midjourney,\footnote{\url{https://www.midjourney.com/}} and Gradio apps \cite{Abid_2019_Gradio}. We envision that a comparable workflow for RL would have a convenient interface for a user to describe their problem, after which a policy---ideally without requiring any further training (see \refsection{Sec:GeneralisationRL})---would be able to start taking actions and solve the problem. In addition to easing the deployment of RL by non-engineers for their tasks of interest, domain experts of novel problems would also become able to create interesting new benchmark domains for RL researchers---similar to how \cite{Eimer_2021_DACBench} aimed to improve accessibility and facilitate further research for dynamic algorithm configuration. This leads to our position as follows:

\begin{tcolorbox}[colframe=orange!75!black,colback=orange!15!white,title=Position]
    The RL research community should place greater focus on benchmarks with environments defined in user-friendly DSLs or natural languages.
\end{tcolorbox}

% \begin{position} \label{Position:position}
%     The RL research community should place greater focus on benchmarks with environments defined in DSLs or natural languages.
% \end{position}

Two clear lines of research that follow from Assumptions~\ref{Assumption:MoreUserFriendly} and~\ref{Assumption:UserFriendlinessGood} would be the design of user-friendly DSLs for relevant application domains, as well as generating reliable translations from natural language to exectuable simulators. However, beyond these challenges related to getting simulators to run in the first place, we also argue that they should be used extensively as benchmarks in general RL research, and that existing benchmarks---with environments implemented directly in general-purpose programming languages---are not sufficient to evaluate how different algorithms and approaches might perform when later applied to environments defined in more user-friendly languages.

For example, consider the common assumption that the full action space $\mathcal{A}$ can be defined in advance, as described in \refsubsection{Subsec:StandardPractice}. While standardised APIs such as OpenAI Gym's \cite{Brockman_2016_OpenAIGym} have undoubtedly aided and accelerated RL research, there is a risk that convergence of the community on such an API may have entrenched this assumption in the community. The ubiquity of this assumption may also be explained by its convenience in the context of deep learning research. There exists some early deep RL work \cite{Riedmiller_2005_Neural,Lange_2010_DeepAutoEncoder} where actions were treated as inputs of neural networks---hence requiring separate forward passes for every legal action to compute policies or state-action values. However, it quickly became common practice---especially after the work on Deep $Q$-Networks by \cite{Mnih_2013_Playing}---to treat actions as outputs. Requiring only a single DNN forward pass per state greatly improves computational efficiency, at the cost of requiring prior knowledge of the full action space. 
%\textcolor{orange}{For the ProcGen framework \cite{Cobbe_2020_ProcGen}, which was designed specifically to challenge and benchmark generalisation in RL with a suite of different environments and procedurally generated levels, the authors designed all the different environments to have the same (15-dimensional, discrete) action space ``to support a unified training pipeline.'' This severely limits the dimensions along which environments may be varied and generalisation of agents can be challenged, and while it can be useful to have a benchmark focused specifically on variation within the state and observation spaces, such a design choice may pessimistically also be viewed as benchmarks being designed to accomodate potential weaknesses of prevailing and popular techniques (DNNs), rather than challenging them.}

In practice, the full action space (or a reasonably sized superset thereof) cannot always be automatically inferred from environment descriptions written in languages that prioritise aspects such as usability over support for robust automated inference. In the relatively verbose, logic-based S-GDL \cite{Love_2008_GDL,Genesereth_2014_GGP}, this is possible, and it is straightforward to build policy networks accordingly \cite{Goldwaser_2020_DeepRLGGP}. In contrast, in Ludii's DSL, which is substantially more succinct and arguably user-friendly \cite{Piette_2020_Ludii}, this does not appear to be feasible. The root of the issue is that succinct descriptions of, for example, game rules, are generally descriptions of procedures that may be used in any game state to generate the set of legal actions for that particular game state. Determining the full action space of the environment requires combinations of this information with an inference of what the entire state space may look like, and this is challenging if the semantics of the DSL are not readily available in a logic-based format. 
For example, the rule that legal moves consist of players placing one of their pieces on any empty cell in the game of \textit{Hex} is formulated as \texttt{(play (move Add (to (sites Empty))))} in Ludii's DSL.
%For example, the rule \texttt{(play (move Add (to (sites Empty))))} describes that legal moves consist of players placing one of their pieces on any empty cell in the game of Hex in Ludii's DSL. 
In combination with knowledge of the size of the board (which is defined in a different rule), knowledge that there are no rules that can ever change the size of the board, and knowledge that there are no other rules for other types of moves, it is easy for humans to infer that the action space of this game must be equal to the number of cells on the board. However, without explicit, direct access to formal semantics of the many hundreds of keywords in Ludii's DSL \cite{Browne_2020_LLR}, there is no clear way to make this inference in an automated and general manner that works for any game described in the language. Practical attempts at using deep learning with Ludii have therefore faced challenges such as \textit{action aliasing}, where a single output node of a policy network may end up getting shared by multiple distinct legal actions \cite{Soemers_2022_DeepLearning}---an issue that is rarely considered possible in other deep RL research. \cite{Maras_2024_Fast} opted to forgo training a policy head altogether, sticking only to a state value function, for games written in another DSL. A similar problem surfaces in PDDLGym \cite{Silver_2020_PDDLGym}, which also requires careful treatment of action spaces due to a mismatch between PDDL and the customary assumptions about action spaces in RL.

These examples of multiple existing DSLs that conflict with the otherwise common assumption of prior knowledge of the full action space is merely one example of an important issue that is largely overlooked by current research. It cannot be ruled out that other types of issues, which are not adequately accounted for by the currently prevailing research methodologies, may surface as the community shifts focus to more benchmarks based on environments defined in DSLs or natural language.

\section{Leveraging Environment Descriptions for Generalisation in RL} \label{Sec:GeneralisationRL}

The previous section argues for the relevance of developing RL techniques that can operate on environments defined in DSLs or natural language, as opposed to general-purpose programming languages, the importance dedicating substantial research efforts on benchmarks following the same methodology, and potential issues that may surface and are underexplored in the current research landscape. However, in addition to potential issues, we also see opportunities. In particular, succinct---but complete---environment descriptions may serve as a powerful tool to improve (zero-shot) generalisation \cite{Kirk_2023_SurveyZeroShot} across the set of all environments that may be described in the language of choice.

\subsection{Generalisation in RL}

The most straightforward setting in RL \cite{Sutton_2018_RL} is to have an agent interacting, collecting experience, and training in a single environment for some time, and to subsequently evaluate its performance in the same environment. This approach has a high risk of producing agents that overfit, in the sense that they may become overly reliant on spurious features, largely ignore state observations altogether and simply memorise trajectories of states or actions, or otherwise be incapable of handling even minor variations on the environment after training \cite{Whiteson_2011_Protecting,Machado_2018_Revisiting,Zhang_2018_Overfitting,Zhang_2020_Dissection}.

A popular category of RL research with a higher degree of generalisation involves training agents on a subset of one or more closely-related environments, and evaluating them in the same set, or a different set of similar environments \cite{Wilson_2007_MultiTask,Li_2009_MultiTask,Fernandez_2013_Learning,BouAmmar_2014_Online,Deisenroth_2014_MultiTask,Farebrother_2018_Generalization,Justesen_2018_Illuminating,Nichol_2018_Gotta,Cobbe_2019_Quantifying,Cobbe_2020_ProcGen,Stone_2021_Distracting,Mediratta_2024_Study}. Different environments in this case may be different levels of the same video game, or subtle variants of an environment with, for example, modified background or foreground colours or patterns, different values for the velocities of certain entities or other numeric parameters, or different reward functions. Particularly in the cases of meta RL \cite{Schmidhuber_1994_MetaLearning,Finn_2017_MAML,Nagabandi_2019_Learning,Rakelly_2019_Efficient,Zintgraf_2022_Fast,Beck_2023_Survey} and few-shot transfer learning \cite{Taylor_2009_TransferRL,Lazaric_2012_TransferRL,MullerBrockhausen_2021_Procedural,Zhu_2023_TransferDeepRL}, at least a small amount of environment-specific fine-tuning prior to evaluation is also assumed. While prior research collectively covers variation along all dimensions of environments (variation in transition dynamics, variation in colours used in state observations, variation in goals or reward functions, etc.), the work described in each publication individually tends to be restricted to a smaller subset of these dimensions. Research on transfer learning in RL \cite{Taylor_2009_TransferRL,Lazaric_2012_TransferRL,Zhu_2023_TransferDeepRL} is often similarly restricted in scope \cite{Glatt_2016_Towards,Rusu_2016_Progressive,Tessler_2017_Hierarchical,Sobol_2018_Visual,Gamrian_2019_Transfer,Mittel_2019_visual,Glatt_2020_DECAF,Yang_2021_learn}. \cite{Soemers_2023_Towards} used DSL-based environment descriptions for (zero-shot) transfer learning between different board games, but only to a relatively small degree, where the transfer mechanism was not automatically learnt. \cite{Banerjee_2007_GeneralGameLearning,Kuhlmann_2007_GraphBased} automatically identified mappings or transferable features between games, but they used a low-level logic-based DSL, which is arguably lacking in terms of user-friendliness.

Sim-to-real transfer \cite{Zhao_2020_SimToReal}---where policies are trained in simulation, but deployed on robots in the physical world---is another major category of generalisation in RL. However, in this case, the need for generalisation is an unfortunate reality due to inevitable differences between simulators and the real world. While some degree of variation is inevitable in this setting, it is typically intentionally kept as low as possible.

\subsection{Leveraging Context for Generalisation}

Theoretical work suggests that, in the worst case, strong assumptions on the similarity between different environments are required for efficient generalisation to be possible: different environments must share an optimal policy \cite{Malik_2021_Generalizable}. One reason for the difficulty of generalisation to unseen environments, without strong restrictions on the degree of variation, is that epistemic uncertainty about relevant parameters of the current environment essentially turns the collection of all environments that the agent may face into a partially observable environment---even if the current state of each individual environment is fully observable \cite{Ghosh_2021_Generalization}. 

The notion of such a collection of environments, each of which may be identified by certain parameters (a \emph{context}), of which some may never be used for training and only appear at test time, may be formalised as a \emph{contextual} \cite{Kirk_2023_SurveyZeroShot,Benjamins_2023_Contextualize} (PO)MDP or Markov game. In the formalism of \cite{Kirk_2023_SurveyZeroShot}, contexts may be as simple as just the value of a random seed that is used for procedural level generation, or take a more complex form such as a vector of parameters that describe important properties of the environment. Contexts may or may not be observable to the agent(s), although the ability to observe contexts---which should also carry sufficient information to enable disambiguation between environments---is required to resolve partial observability \cite{Ghosh_2021_Generalization} induced by epistemic uncertainty.

Research on multi-task RL often involves providing contexts as inputs to agents, but these contexts tend to be far from full environment descriptions. For example, \cite{Deisenroth_2014_MultiTask} provide a single or a handful of numeric values as context, which describe goal coordinates for robotic control tasks. It is common to provide short, language-based instructions or hints to guide the agent \cite{Maclin_1996_Creating,Andreas_2017_Modular,Misra_2017_Mapping,Oh_2017_ZeroShot,Shu_2018_Hierarchical,Williams_2018_Learning,Luketina_2019_Survey,Goyal_2021_PixL2R,Lifschitz_2023_Steve1,Kharyal_2024_GLIDERL}, but such instructions do not (fully) describe the environment. In cases where goals correspond to elements of the state space, universal value function approximators \cite{Schaul_2015_UVFA} can learn reusable skills and generalise by conditioning on goal states. \cite{Lee_2023_Supervised,Raparthy_2023_Generalization,Reed_2023_Generalist} prompt agents with demonstrations of interactions by experts for disambiguation between environments and in-context learning, which is a form of context that is arguably more difficult to acquire than environment descriptions (requiring an environment-specific expert to have already been trained), whilst simultaneously carrying less information (it does not reveal information about any parts of the environment that are not explored in the demonstration). CARL \cite{Benjamins_2023_Contextualize} extends several well-established RL benchmarks with contexts, but these contexts take the form of dictionaries specifying values for only a handful of variables. \cite{Rostami_2020_TaskDescriptions} consider more elaborate task descriptions, but not necessarily ones that fully specify complete environments. \cite{Sun_2020_Program} use a DSL to prescribe policies that an agent should execute, as opposed to describing the environment itself. \cite{Branavan_2012_Learning} leverage text from strategy guides to guide search-based game playing agents, but these texts do not fully describe the environment, and were not used to improve generalisation to unseen environments. The textual descriptions provided to agents by \cite{Zhong_2020_RTFM} are perhaps closest to what we propose, although their descriptions are not sufficiently detailed to the extent that they could be compiled into a correct simulator, and are not meant to serve as a substitute for implementing the environment in a programming language.

\subsection{Environment Descriptions as Context}

If it is desirable to describe environments in succinct DSLs or in natural language, as posited in \refsection{Sec:DescriptionLanguagesEnvironments}, then these descriptions may also be used to improve generalisation by serving as contexts. Crucially, leveraging such descriptions as context should not be viewed by peer reviewers as being a reduction in generality, or being restricted to a particular DSL, as the general workflow of providing environments in such a language is arguably more accessible and more general than using a programming language for many potential end users. An important property of such environment descriptions is that they come from a shared language, and it should be possible for humans as well as software to generate novel environment descriptions in the same language. We cannot only generate contexts from environments, but also generate environments (in the form of fully executable simulators) from contexts. From the researchers' point of view, this is valuable as it makes environments easily controllable and enables a wide variety of evaluation protocols \cite{Kirk_2023_SurveyZeroShot}. From the learning agent's point of view, this property may also be valuable in that a program could actively learn about the description language that is used by procedurally generating new descriptions \cite{Browne_2009_PhD,Todd_2024_GAVEL}, translating them into executable simulators, and learning in them---effectively forming their own curriculum of environments to learn from \cite{Dennis_2020_Emergent,ParkerHolder_2022_Evolving,Jackson_2023_Discovering,Samvelyan_2023_MAESTRO,Rigter_2024_RewardFree}.

Furthermore, it could be argued that contexts that completely describe an environment---to the extent that they could be translated into executable simulators---are likely to be a prerequisite for unrestricted, zero-shot generalisation in RL \cite{Irpan_2019_Principle}. Consider the generalisation abilities of humans. In some cases, humans can effectively generalise to unseen situations without relying on explicit task descriptions, but in others they cannot. For example, if a human plays a new video game for the first time, in which there is something that looks like fire, they can infer that they should likely avoid the fire---based on their related experience in the physical world and other video games. This relies on an implicit assumption that fire in this particular video game works similarly to how it works in other games, or in the real world, which may be incorrect. If a human is faced with a brand new board game, they cannot be expected to play it well if the rules of the game are not explained. Once the rules are explained, they may be able to play well immediately---based on their experience with related board games and ability to reason---without any direct experience with the game in question.

\section{Other Barriers to Widespread Deployment of RL} \label{Sec:OtherBarriers}

While the focus of this paper is on the challenge posed to widespread adoption of RL by the customary use of programming languages for defining environments or tasks, this is not the only barrier. The choice to focus on environment descriptions and the languages in which they are described is due to our perception that this aspect of RL research and deployment is hardly if ever discussed, whereas other challenges are already more frequently acknowledged and discussed. However, we do not expect these different challenges, or solutions to them, to be completely orthogonal, and discuss how they may be related here.

\subsection{Sample Efficiency and Computation Resource Requirements} \label{Subsec:SampleEfficiency}

Sample (in)efficiency and the requirement for substantial computation resources are frequently cited as pervasive issues in deep RL, both in terms of feasibility of deployment in the ``real world'' as well as research more generally \cite{ObandoCeron_2021_Revisiting,Agarwal_2022_Reincarnating,Mohan_2024_Structure}. This issue motivates the recent trend of implementing simulators to be entirely runnable on hardware accelerators via frameworks such as JAX, as discussed in \refsubsection{Subsec:StandardPractice}. When applicable, such an approach is highly effective at lowering the barrier to entry in terms of hardware requirements for RL research. However, this style of implementing environments is arguably even more demanding in terms of engineering expertise than using general-purpose programming languages, and is not directly helpful in terms of lowering the barrier to entry for deployment of RL by users without such engineering expertise.

In contrast, environments implemented in commonly-used DSLs \cite{Love_2008_GDL,Schaul_2014_Extensible,Piette_2020_Ludii} tend to have substantially lower simulation speeds than similar environments implemented directly in programming languages like \texttt{Java} or \texttt{C++}. This only exacerbates the need for computation resources---to make up for low simulation speeds---if RL is to be successful in these environments. We do not view this as an argument against using DSLs, but rather as an additional reason to focus even more on improving sample efficiency of RL. Only focusing on (methodologies for implementing) environments that permit running simulations at high speeds would be an inadvisable form of tunnel vision, not unlike how the \textit{hardware lottery} \cite{Hooker_2021_Hardware} describes when certain algorithms receive more attention than others depending on their compatibility with current hardware. Alternatively, it may be possible to combine the high speed of hardware-accelerated programming with user-friendly domain specific languages. The Ludax framework \cite{Todd_2025_Ludax} is a recent, and to our knowledge first, example of a system that compiles user-friendly game description languages (inspired by Ludii's language \cite{Piette_2020_Ludii}) into native and hardware-accelerated simulation code via JAX. To what extent this can be effectively replicated with other languages, domains other than board games, or even simply a broader and more diverse set of board games than that currently supported by Ludax, remains to be seen in future research.

\subsection{Complexity of Agent Design, RL Algorithm Selection, and Hyperparameter Tuning} \label{Subsec:AutoRLMetaRL}

Another challenge for the deployment of RL is lack of reliability (e.g., instability or high variance in performance levels over different random seeds), and the degree to which RL performance depends on selecting the right algorithms, optimisers, neural network architectures, hyperparameters, and other design decisions for each specific environment \cite{Andrychowicz_2021_WhatMatters,Eimer_2023_Hyperparameters,ObandoCeron_2024_Consistency}. All of these are choices that tend to require a substantial amount of RL and engineering expertise, or the ability to run extensive experiments (e.g., grid searches for hyperparameters). However, advances in meta RL \cite{Schmidhuber_1994_MetaLearning,Finn_2017_MAML,Nagabandi_2019_Learning,Rakelly_2019_Efficient,Zintgraf_2022_Fast,Beck_2023_Survey} and AutoML (or AutoRL) \cite{ParkerHolder_2022_Automated,Mohan_2023_AutoRL,Goldie_2024_CanLearned} may help lower this barrier to entry \cite{Lindauer_2024_Position}. It is plausible that the ability to provide succinct and information-rich environment descriptions as context, as discussed in \refsection{Sec:GeneralisationRL}, may also benefit these techniques.

\section{Proposed Research Agenda} \label{Sec:ProposedResearchAgenda}

Building on the arguments and analyses from the previous sections, we propose a research agenda centred around the use of user-friendly languages to describe problems of interest (i.e., the environments in RL settings), with anticipated opportunities, new avenues for research, and challenges regarding aspects such as democratisation of the use of AI (specifically, RL), description language design, generalisation and transfer, sample efficiency, and AutoRL and meta RL. 

\begin{figure}[t]
\centerline{\includegraphics[width=\textwidth]{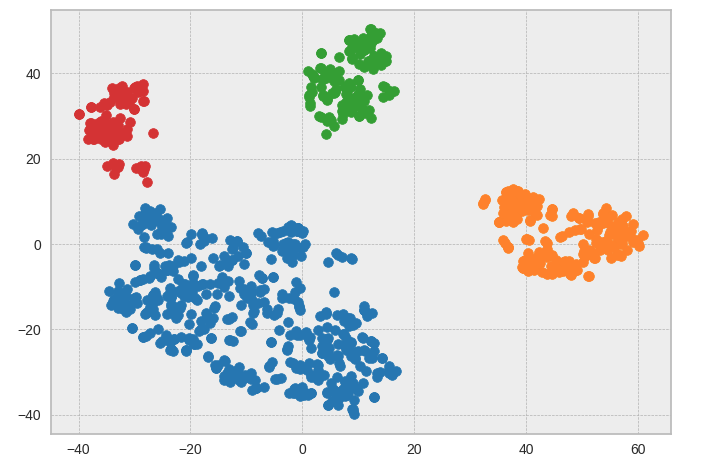}}
\caption{A set of 1059 different board games, described in Ludii's game description language, reduced from a larger feature space \cite{Piette_2021_Concepts} to two dimensions via a t-SNE embedding \cite{Maaten_2008_TSNE}. Image source: \cite{Stephenson_2023_Measuring}.}
\label{Fig:GameClusters}
\end{figure}

While it is not necessarily an exhaustive list, and relative importance of each item may vary among different research and deployment contexts, we identify the following list of desiderata for environment description languages:
\begin{itemize}
    \item \emph{User-friendliness}: enabling end users with little programming or RL expertise to describe their optimisation problems of interest, such that RL can be applied to tackle them, is central to our proposed research agenda. This implies that any languages used to these environments ought to be user-friendly.
    \item \emph{Complete and unambiguous environment specifications}: there is a substantial amount of work on providing \textit{some} information about an environment as context to aid decision-making \cite{Maclin_1996_Creating,Andreas_2017_Modular,Misra_2017_Mapping,Oh_2017_ZeroShot,Shu_2018_Hierarchical,Williams_2018_Learning,Luketina_2019_Survey,Rostami_2020_TaskDescriptions,Sun_2020_Program,Zhong_2020_RTFM,Goyal_2021_PixL2R,Benjamins_2023_Contextualize,Lifschitz_2023_Steve1,Kharyal_2024_GLIDERL}, but unless descriptions are complete to the extent that they can be compiled into an executable simulator, they do not help alleviate the need for engineers with programming and RL expertise. A key exception is when training will be done in the real world (e.g., directly on physical robots), in which case only specifying goals, objectives, or reward functions would suffice.
    \item \emph{Generality and expressiveness}: we remark that, as the orange boxes in \reffigure{Fig:DSLEnvironment} indicate, we still expect programming and RL expertise to be required in some parts of the proposed workflow. In particular, such expertise will still be required to build compilers for any designed language. Therefore, this workflow will only pay off, in terms of democratisation of the use of RL, if the same compiler can be re-used for multiple different (but presumably related) tasks, all describable without engineering expertise in the same user-friendly language. If there is only a single task of interest, a simulator for it might as well be implemented directly in code. This suggests that a sufficient degree of generality and expressiveness of the language is crucial: we ought to be able to describe a variety of different tasks in the same language. In practice, we expect that there may be tension between the desiderata of user-friendliness and generality. While we have no quantitative evidence at this point, our own practical experience, and conversations with others among the general game playing research community in particular, suggest that description languages that specifically target a restricted subset of domains such as video games \cite{Schaul_2013_VGDL,Schaul_2014_Extensible} or board games \cite{Piette_2020_Ludii}, using high-level keywords specific to that domain, tend to be easier to use than lower-level (e.g., logic-based) languages \cite{Love_2008_GDL,Genesereth_2014_GGP}.
    \item \emph{Computational efficiency}: as discussed in \refsubsection{Subsec:SampleEfficiency}, high simulation speeds are desirable to speed up training, in particular due to the lack of sample efficiency in current RL techniques. Therefore, descriptions can ideally be compiled into computationally efficient simulators.
    \item \emph{Facilitation of procedural generation}: especially in the case of highly expressive languages, it may be infeasible to manually write enough descriptions to provide sufficient coverage of the full space of all describable problems, which can in turn impair the ability to train policies that effectively generalise in a zero-shot manner across the space. For example, the t-SNE visualisation \cite{Maaten_2008_TSNE} in \reffigure{Fig:GameClusters} depicts how even a relatively large set of 1059 different board games still leaves large uncovered areas in the underlying space, which may be important to fill up with further example games for a model to effectively learn the precise semantics of the game description language, and implications on effective playing strategy. This may be addressed by procedurally generating new tasks to fill up uncovered parts of the space \cite{Todd_2024_GAVEL}. The design of the language and underlying problem representation can have a substantial impact on how easy or difficult it is for, e.g., evolutionary algorithms to navigate and construct suitable, novel elements of this space.
\end{itemize}

\noindent All of these desiderata should feature in evaluations of designed languages in our proposed research agenda. For some aspects, such as measuring computational efficiency of frameworks \cite{Kowalski_2020_Efficient,Todd_2025_Ludax} and (theoretically) analysing the expressiveness of environment and goal description languages \cite{Thielscher_2011_Universal,Kowalski_2019_Regular,Lin_2022_Expressive,Soemers_2024_Universal,Sakcak_2025_Limits}, this is already commonplace. Other aspects, such as ease of use for humans or suitability of representations for procedural content generation, are sometimes listed as design considerations \cite{Piette_2020_Ludii}, but lacking in terms of quantitative and objective evaluations in existing research (though there is some recent work in such directions \cite{Rani_2025_GoalsVsRewards}). 

When we have a user-friendly description language in which end users can conveniently describe a variety of problems of interest, we still require an RL solution to tackling these problems. For the same end users to be able to also execute this step, we expect to require substantial advances in (i) generalisation, and/or (ii) AutoRL and meta RL. Advances in generalisation (discussed in more detail in \refsection{Sec:GeneralisationRL}) would enable us to pre-train an agent in a wide variety of environments described in the chosen language, and have it automatically generalise to novel problems described in the same language. In the case of successful zero-shot generalisation, this could even circumvent the need for training time and other resources (e.g., hardware) for the end user. Similarly, advances in AutoRL and meta RL (discussed in more detail in \refsubsection{Subsec:AutoRLMetaRL}) would circumvent the need for the end user to have the RL knowledge necessary to choose which algorithm(s) to use, but would not alleviate the need for training resources such as time and hardware.

\section{Related Work} \label{Sec:RelatedWork}
Mannor and Tamar \cite{Mannor_2023_Towards} also caution against the excessive focus of the research community on algorithms in existing benchmarks, with little attention for deploying to novel problems, but they do not discuss ease of use, or user-friendly environment description languages as a potential solution. Zhu et al. \cite{Zhu_2023_Pearl} present Pearl as a framework meant to facilitate users applying RL to their real-world applications. However, this framework focuses on the design of agents, as opposed to environments, and does not alleviate the need for highly engineered environment implementations. 

Rodriguez-Sanchez et al. \cite{Rodriguez_2023_RLang} introduce RLang as a DSL that can be used to provide background or expert knowledge on any aspect of an MDP. However, they propose for such descriptions to be provided \emph{in addition to} environment implementations in general-purpose programming languages, rather than as a replacement. Nevertheless, this could be an example of a DSL that could be used for our proposed research agenda. Jothimurugan et al. \cite{Jothimurugan_Composable_2019} describe a DSL that can be used to specify reward functions via, e.g., goals and constraints, but no other aspects of the environment. In the specific case of robotics in the physical world, natural language-based instructions for robots to follow \cite{Ahn_2024_AutoRT} may suffice to provide widespread, easy access, as there is no need for a simulator, but this does not extend to many other (virtual) applications. Focusing specifically on the problem of representing goals (rather than full MDPs or RL problems), and not necessarily from the perspective of users who are not engineering or RL experts, Davidson et al. also consider goal representations \cite{Davidson_2024_Toward} based on programs (essentially DSLs) \cite{Davidson_2024_Goals} and natural languages, among other solutions.

While our position and proposed research agenda revolve around ideas such as user-friendliness and democratisation of the use of RL, we still view these issues largely from a technical angle by exploring how we may technically develop techniques to facilitate broader use of RL. Other recent position papers explore related ideas, such as the inherent value of application-driven AI research, recommended changes to peer review, hiring, and teaching policies in the machine learning community, and legal and ethical considerations when shifting research focus to real-world applications \cite{Rolnick_2024_ApplicationDriven,Hartman_2025_ResponsibleApplicationDriven}. In another recent position paper, Blili-Hamelin et al. \cite{BlaliHamelin_2025_AGINorthStar} argue to shift away from an excessive focus on Artificial General Intelligence as a research goal, and rather be more specific and explicit about research goals and the exact meaning of notions such as ``generality.'' Our proposed research agenda is in line with this, as we argue to focus on restricted subsets of problems, and build agents that can generalise specifically within the set of problems describable in any description language of choice.

\section{Conclusion} \label{Sec:Conclusion}

The central position in this paper is to argue for a need for increased focus in the reinforcement learning (RL) research community on benchmarks in which environments are not implemented directly in general-purpose programming languages, but rather described in user-friendly domain-specific languages (DSLs) or even natural languages. The core reason for this is to empower end users with little to no programming or RL expertise to apply RL as a solution to their optimisation problems of interest: they only need to be able to describe their problems in the user-friendly language of choice. Furthermore, we see potential for advances in (zero-shot) generalisation and transfer within the set of problems describable in any such language, by using the environment descriptions as context on which to condition policies, value functions, and so on. We presented a complete outline of the research agenda we propose, accounting for desiderata and challenges such as user-friendliness, sample efficiency, and generalisation.

\begin{credits}
% \subsubsection{\ackname} A bold run-in heading in small font size at the end of the paper is used for general acknowledgments, for example: This study was funded by X (grant number Y).

\subsubsection{\discintname}
The authors have no competing interests to declare that are relevant to the content of this article.
\end{credits}
%
% ---- Bibliography ----
%
% BibTeX users should specify bibliography style 'splncs04'.
% References will then be sorted and formatted in the correct style.
%

\bibliographystyle{splncs04}
\bibliography{Dennis-Soemers-Bib}

\end{document}